\documentclass[10pt,twocolumn,letterpaper]{article}

\usepackage{cvpr}
\usepackage{times}
\usepackage{epsfig}
\usepackage{graphicx}
\usepackage{amsmath}
\usepackage{amssymb}
\usepackage{bm}

\usepackage[breaklinks=true,bookmarks=false]{hyperref}

\cvprfinalcopy 

\def\cvprPaperID{2984} 

\ifcvprfinal\fi
\begin{document}

\title{Multi-source weak supervision for saliency detection}

\author{Yu Zeng$^1$, Yunzhi Zhuge$^1$, Huchuan Lu$^1$, Lihe Zhang$^1$, Mingyang Qian$^1$, Yizhou Yu$^2$\\
$^1$ Dalian University of Technology, China\\
$^2$ Deepwise AI Lab, China\\
{\tt\small zengyu@mail.dlut.edu.cn, zgyz@mail.dlut.edu.cn, lhchuan@dlut.edu.cn, }\\
{\tt\small zhanglihe@dlut.edu.cn, mingyangqian25@gmail.com, yizhouy@acm.org}
}

\maketitle
\thispagestyle{empty}

\begin{abstract}
The high cost of pixel-level annotations makes it appealing to train saliency detection models with weak supervision. However, a single weak supervision source usually does not contain enough information to train a well-performing model. To this end, we propose a unified framework to train saliency detection models with diverse weak supervision sources. In this paper, we use category labels, captions, and unlabelled data for training, yet other supervision sources can also be plugged into this flexible framework. We design a classification network (CNet) and a caption generation network (PNet), which learn to predict object categories and generate captions, respectively, meanwhile highlight the most important regions for corresponding tasks. An attention transfer loss is designed to transmit supervision signal between networks, such that the network designed to be trained with one supervision source can benefit from another. An attention coherence loss is defined on unlabelled data to encourage the networks to detect generally salient regions instead of task-specific regions. We use CNet and PNet to generate pixel-level pseudo labels to train a saliency prediction network (SNet). During the testing phases, we only need SNet to predict saliency maps. Experiments demonstrate the performance of our method compares favourably against unsupervised and weakly supervised methods and even some supervised methods.
\end{abstract}

\section{Introduction}
Saliency detection aims to detect the most informative parts of an image. It can be applied to benefit a wide range of applications~\cite{avidan2007seam, craye2016environment, zund2013content}, and thus has attracted a lot of interest in recent years. Driven by the remarkable success of deep convolutional neural networks (CNNs), a lot of attempts have been made to train CNNs for saliency detection~\cite{hou2017deeply, liu2016dhsnet, wang2016saliency, wang2015deep}. CNN-based methods usually need a large amount of data with pixel-level annotations for training. Since it is expensive to annotate images with pixel-level ground-truth, attempts have been made to exploit higher-level supervision, \eg image-level supervision, to train CNNs for saliency detection~\cite{wang2017learning}. 
\begin{figure}[t]
\begin{center}
  \includegraphics[width=.9\linewidth]{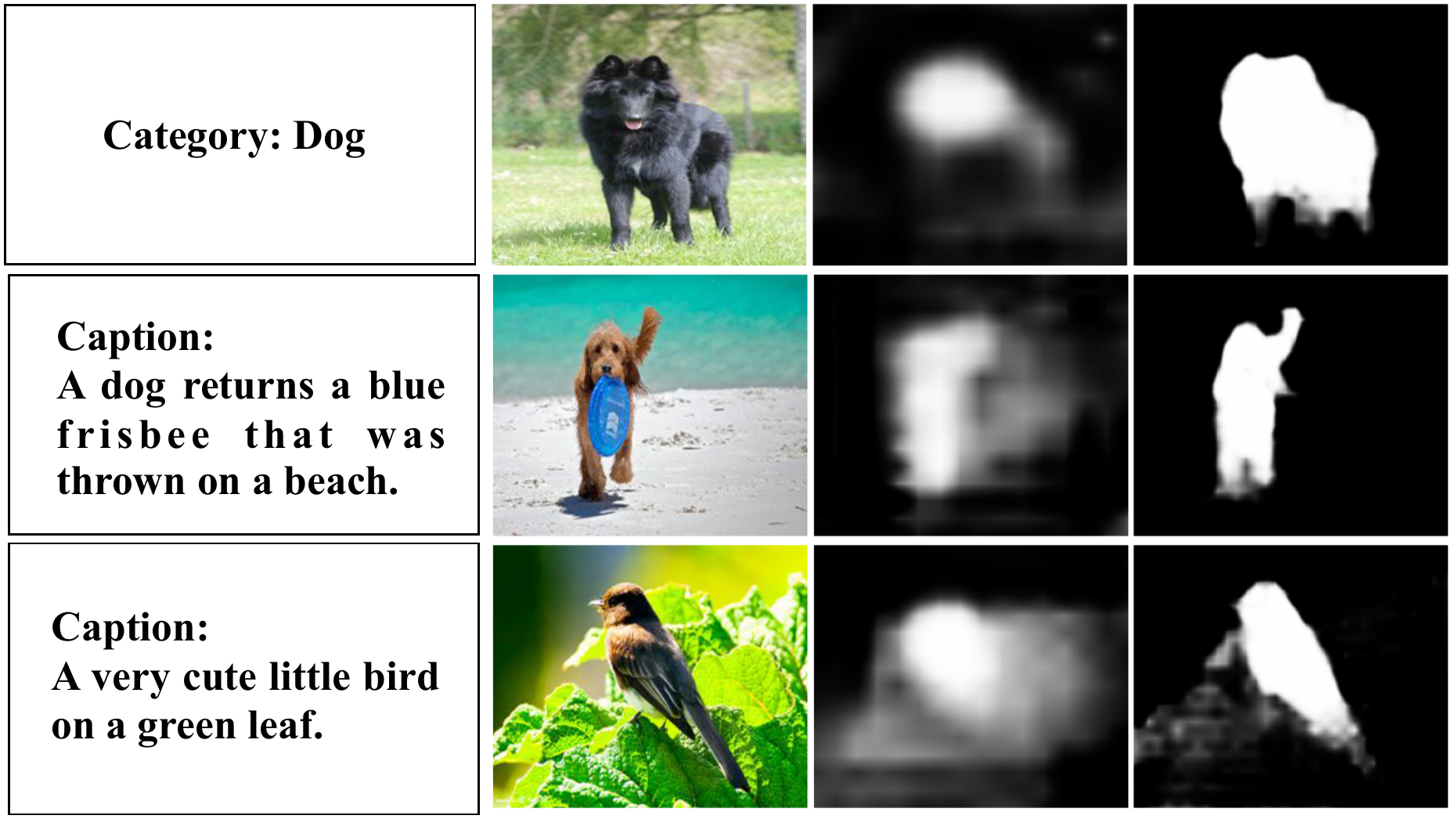}\\
  (a)~~~~~~~~~~~~~~~~~~~(b)~~~~~~~~~~~~~~~~(c)~~~~~~~~~~~~~~~~(d)
\end{center}
  \caption{(a) annotations. (b) images. (c) saliency maps of the models trained with single weak supervision source shown in the first column of the corresponding row. (d) saliency maps of the models trained with our proposed multi-source weak supervision framework. }
\label{fig1}
\vspace{-10pt}
\end{figure}

However, it is challenging to train a network to cut the salient objects accurately in weak supervision settings. On the one hand, weak supervision sources are incomplete and noisy. For example, the image-level category label is an efficient weak supervision cue for saliency detection. It indicates the category of the principal objects in it, which are much likely to be the salient foreground. However, category labels are too simple to convey sufficient information. Without knowing the attribute or motion of a salient object, the network trained with category tags might only highlight the most discriminative region instead of the whole object. As shown in the first row of Figure~\ref{fig1}, the model trained with category labels only highlights the face of the dog as the face provides enough information to categorize it as a dog. 
Another weak supervision cue is the image caption. Image captions are a few sentences that describe the main content of an image. Compared with image-level tags, captions provide more comprehensive descriptions of the salient objects. As shown in the second row of Figure~\ref{fig1}, for a picture of a dog, the caption not only tells there is a dog but also says that the dog is returning and is with a frisbee. To generate the correct caption, the network needs to attend the whole dog. Therefore, the network trained with captions is more likely to capture the entire salient objects. However, image captions usually describe not only the salient objects but also the background. This might lead to inaccurate saliency detection results. As shown in the second and the third rows of Figure~\ref{fig1}, apart from the salient objects such as the bird and the dog, the captions also mention the background keywords such as the beach and the green leaf. As a result, saliency maps of the networks trained with captions highlight a part of the background. 

On the other hand, although it is appealing to integrate multiple weak supervision sources due to their complementarity, there are still plenty of obstacles to it. 
First, there is a lack of large scale dataset with multiple kinds of annotations, while the existing datasets with different annotations are unmatched for saliency detection task.
Second, the models trained by using different annotations are usually required to have different structures. Therefore, it is worth designing a unified framework to combine these models and benefit from multiple sources of annotations. 

\begin{figure*}[t!]
\begin{center}
\includegraphics[width=.9\linewidth]{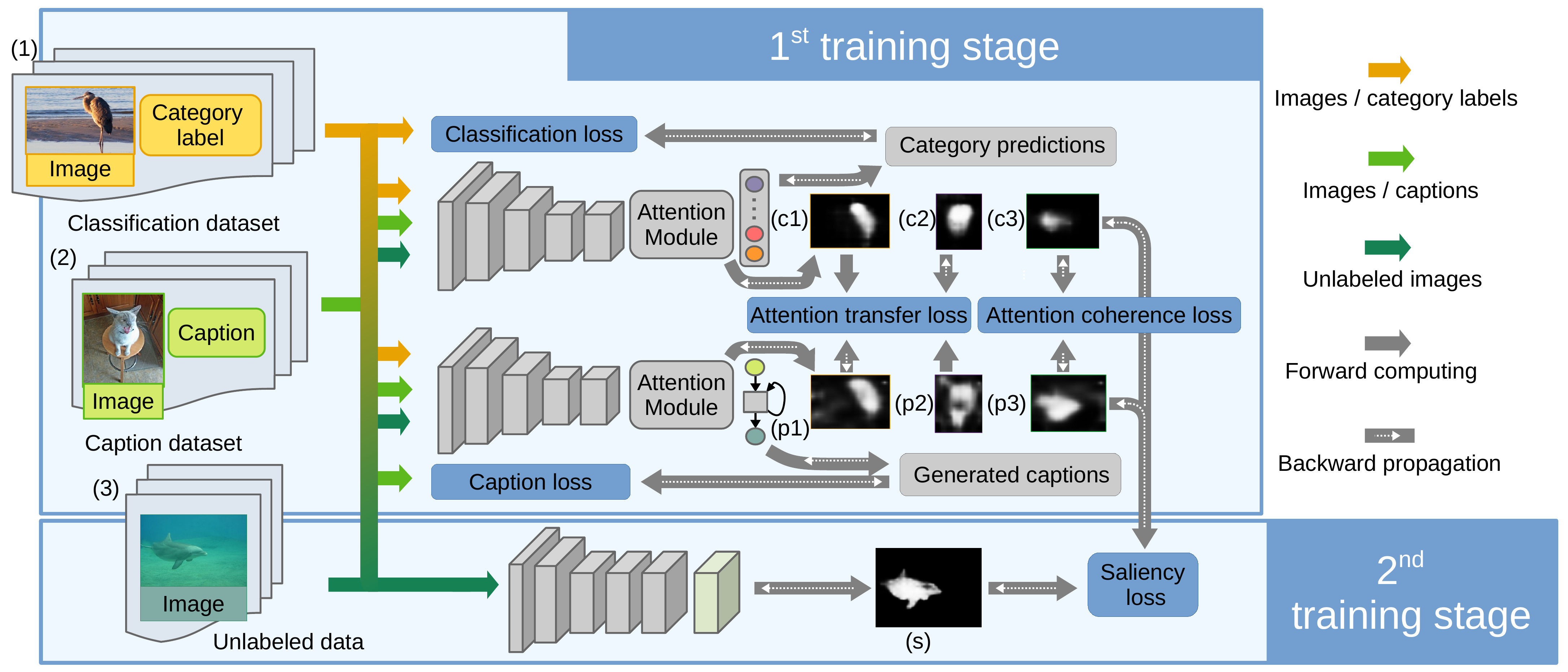}
\end{center}
  \caption{An overview of the proposed multi-source weak supervision framework. (1, 2, 3) images annotated with category labels, caption annotations, and unlabelled images. (c1, c2, c3) saliency maps of images (1, 2, 3) generated by the classification network (CNet). (p1, p2, p3) saliency maps of images (1, 2, 3) generated by the caption generation network (PNet). (s) Final output saliency maps. }
\label{fig2}
\end{figure*}

To this end, we propose a weakly supervised learning framework that integrates multiple weak supervision cues to detect salient objects. Specifically, we use data annotated with image-level tags, image captions, and unlabelled data. Note that other supervision sources also can be plugged into this flexible framework. We design three subnetworks: a multi-label classification network (CNet), a caption generation network (PNet) and a saliency prediction network (SNet). Figure~\ref{fig2} shows the main architecture. CNet is composed of a convolutional feature extractor, an attention module and a fully connected layer. For an input image, the feature extractor produces a feature vector for each region. 
The attention module generates spatial attention over all regions that control the information flow from each region to the fully connected layer. It has to attend the most important regions to predict the category labels correctly. The spatial attention of all image regions composes a coarse saliency map that highlights all potential category-agnostic object regions. 
%
PNet has a similar structure to CNet, with the fully connected layer replaced by an LSTM~\cite{hochreiter1997long} layer to generate captions. The coarse saliency map generated by its attention module highlights the essential regions for generating correct captions. 

To make full use of the annotations, we design an attention transfer loss to transmit supervision signal between networks, such that the network purposed for one supervision source can benefit from another source. When trained with category labels, CNet learns from the annotations, and PNet learns from the coarse saliency maps of CNet with the attention transfer loss. When trained with the images annotated with captions, PNet learns from the annotation, and CNet learns from the coarse saliency maps of PNet. 
To encourage the networks to detect generally salient regions instead of task-specific regions, we define an attention coherence loss that uses unlabelled data for regularization. The coarse saliency maps of the unlabelled images produced by CNet and PNet are refined according to low-level color similarity, and then coarse saliency maps by CNet and PNet are matched to the refined one. 

After CNet and PNet are trained, we use them to generate pseudo labels to train the saliency prediction network (SNet). 
%
SNet consists of a feature extractor and several convolution layers. Inspired by~\cite{chen2018deeplab}, we use dilated convolution to enlarge the receptive fields and use parallel dilated convolutional layers with different dilation rates to capture objects and context at multiple scales. When testing, we only need SNet to generate the final saliency maps. As shown in the last column of Figure~\ref{fig1}, our proposed multi-source supervision framework can leverage the complementary strengths of diverse supervision sources to generate better saliency maps that evenly highlight the generally salient objects meanwhile suppress the background. 

In summary, our main contributions are as follow:
\begin{itemize}
  \item We propose a novel weak supervision framework to train saliency detection models with diverse supervision sources. As far as we know, this is the first attempt to integrate multiple supervision cues into a unified framework for saliency detection. 
  \item We design three networks for saliency detection that learn from category labels, captions and noisy pseudo labels, respectively.
  \item We propose an attention transfer loss to transmit supervision signal between networks to let the network designed for one supervision source benefit from another source, and an attention coherence loss to encourage the networks to detect the generally salient regions.
  \end{itemize}

\section{Related work}
\subsection{Salient objects detection}
Early research for saliency detection focused on hand-crafted features and heuristic priors~\eg, centre prior~\cite{jiang2013submodular} and boundary background prior~\cite{yang2013saliency}. Recently, driven by the remarkable success of deep convolutional neural networks (CNNs) on various vision tasks, a lot of deep learning based methods have been proposed for saliency detection. Li~\etal~\cite{li2015visual} extracted multi-scale features from a deep CNN to represent superpixels and used a classifier network to predict the saliency score of each superpixel. Hou~\etal~\cite{hou2017deeply} proposed a skip-layer structure with deep supervision for saliency detection. Wang~\etal~\cite{wang2018detect} proposed a global Recurrent Localization Network to exploit contextual information by the weighted response map to localize salient objects more accurately. Although these methods achieved superior performance, they all needed expensive pixel-level annotations for training. 

\subsection{Weakly supervised learning}
To reduce the cost of hand-labelling, weakly supervised learning has attracted increasingly more attention. Pinheiro and Collobert~\cite{pinheiro2015image} used a segmentation network to predict the pixel-level labels and aggregated them into image-level ones. Then the error between the predictions and the image-level ground truth was backpropagated to update the network. Ahn and Kwak~\cite{Ahn_2018_CVPR} utilized class activation maps (CAM)~\cite{zhou2016learning} to train a network to predict semantic affinities within local image areas, which were incorporated with the random walk to revise the CAM and generate the segmentation labels. 
Wang~\etal~\cite{wang2017learning} trained a CNN to detect salient objects with image-level supervision. They designed a Foreground Inference Net (FIN) to inference potential foreground regions and proposed a global smooth pooling (GSP) operation to aggregate responses of the inferred foreground objects. Unlike global max pooling (GMP) and global average pooling (GAP) that perform hard selections of latent instances, GSP explicitly computes the weight of each instance and is better suited to pixel-level tasks. However, GSP needs to solve a maximization problem for each input image, which greatly slows down the forward computation of the network. In contrast, our proposed attention module aggregates the features and compute the spatial distribution of foreground objects in one forward pass, bringing much less computation burden. Moreover, all of the above methods rely on a single image-level supervision source, while we integrate complementary supervision cues to train a more robust model. 

\section{The proposed method}
In this section, we elaborate on the proposed multi-source weak supervision for saliency detection. The overall framework is illustrated in Figure~\ref{fig2}. The classification network (CNet) predicts category labels meanwhile its attention module generates a coarse saliency map that highlights the regions related to the classification results. The caption generation network (PNet) generates captions and locates the corresponding regions. When training with category labels, the category localization loss is computed for CNet and the attention transfer loss is computed for PNet. When training with captions, the caption localization loss and the attention transfer loss is computed for PNet and CNet respectively. When training with unlabelled data, we compute attention coherence loss using saliency maps of CNet and PNet. After CNet and PNet are trained, we use them to generate pseudo labels to train the saliency prediction network (SNet). The architectures of CNet, PNet as well as the saliency prediction network (SNet) are presented in Section~\ref{basenet},~\ref{attention},~\ref{netarch}. The training strategy is described in Section~\ref{train}. 

\subsection{Feature extractors}\label{basenet}
Feature extractors of our networks are designed based on DenseNet-169~\cite{huang2017densely}, which consists of five convolutional blocks for feature extracting and a fully connected linear classifier. Each layer is connected to every other layer within the same block. Owing to its dense connectivity pattern, DenseNet can achieve comparable classification accuracy with a smaller number of parameters than other architectures. 
We remove the fully connected classifier and use the convolutional blocks as our feature extractors. To obtain larger feature maps, we remove the downsampling operator from the last few pooling layers. For CNet and PNet, we only do this pruning in the last pooling layer to make the generated feature maps of $\frac{1}{16}$ the input image size. For SNet, we modify the last two pooling layers to obtain feature maps with more detail information and generate better saliency maps. Feature extractor of SNet outputs feature maps of $\frac{1}{8}$ the input image size.

\subsection{Attention module}\label{attention}
We design an attention module to compute the spatial distribution of foreground objects over the image regions meanwhile aggregate the feature of all regions. Given an input image, the feature extractor generates a feature map denoted as a set of feature vectors $\{\bm{v}_1, ..., \bm{v}_K\}$, each of which encodes an image region (~\ie a spatial grid location in the last convolutional layer of the feature extractor). $K$ denotes the number of regions and $K=H\times W$ for a feature map of spatial size $H\times W$. We apply a $1\times 1$ convolution followed by a sigmoid function on the feature map to generate a coarse saliency map as follow,
\begin{equation}
s_i = \sigma(\bm{w}_s^\intercal \bm{v}_i + b_s), 
\end{equation}
in which $\sigma$ denotes the sigmoid function. $\bm{w}_s$ and $b_s$ are the learned parameters. $s_i$ is the saliency score of the $i$-th region. Saliency scores of all regions constitute a saliency map $S$. 

Given the feature vector $\bm{v}_i$ and saliency score $s_i$ of each region, we compute its attended feature of each region, denoted as $\bm{f}_i$ as follow,
\begin{equation}\label{saliency_of_each_region}
\bm{f}_i = s_i \cdot (\bm{w}_f^\intercal \bm{v}_i + b_f), 
\end{equation}
in which $\bm{w}_f$ and $b_f$ are the learned parameters. This can be implemented by another $1\times 1$ convolutional layer of which the output is multiplied with $S$ element-wise. Then we compute a normalized attention weight $a_i$ for each image region as follow,
\begin{equation}\label{att_weight}
\begin{array}{r@{}l}
a_i &{}= \bm{w}_a^\intercal \bm{f}_i + b_a\\
\bm{\alpha} &{}= \mbox{softmax}(\bm{a}), 
\end{array}
\end{equation}
where each element $a_i$ of the vector $\bm{a}$ is the attention weight of the $i$-th region. $\bm{w}_a$ and $b_a$ are the learned parameters. The softmax function is to constrain the sum of the weight of all positions to 1. 

\begin{figure}[htbp]
\begin{center}
  \includegraphics[width=\linewidth]{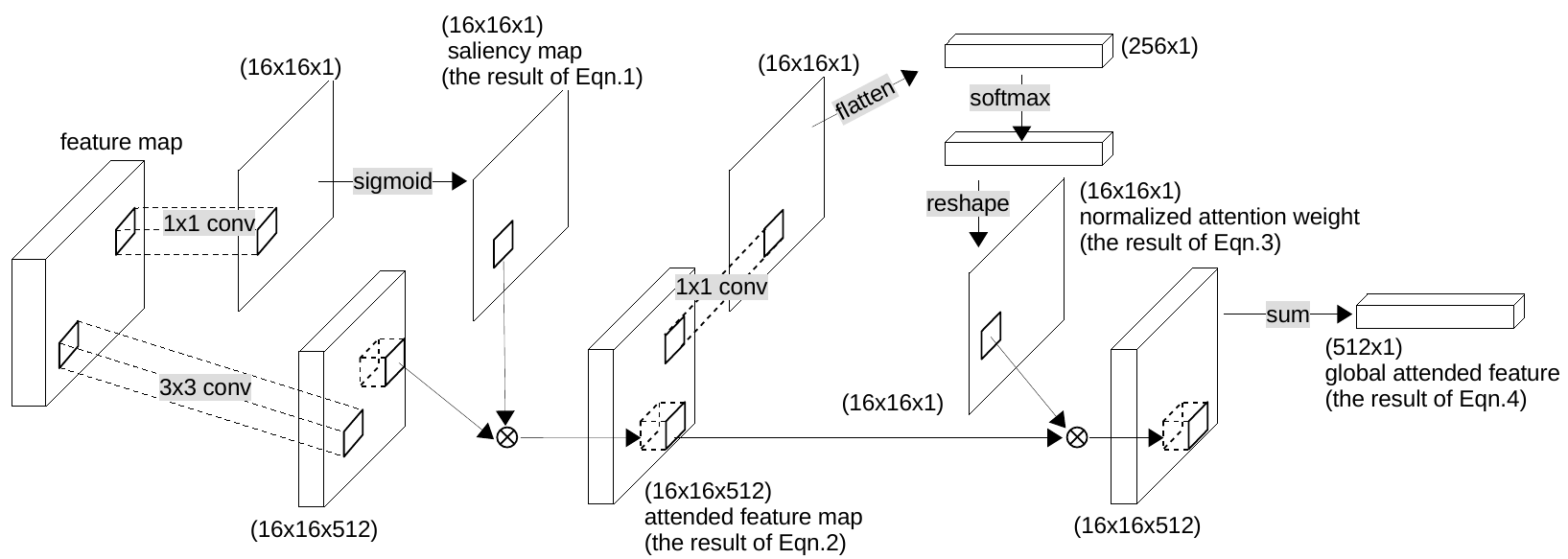}\\
  (a)~~~~~~~~~~~~~~~~~~~(b)~~~~~~~~~~~~~~~~(c)~~~~~~~~~~~~~~~~(d)
\end{center}
  \caption{The details of the attention module.}
\label{att}
\vspace{-10pt}
\end{figure}

Let $\alpha_i$ be the element of $\bm{\alpha}$; the global attended feature $\bm{g}$ of the input image is the weighted average of the attended features of all regions as follow,
\begin{equation}
\bm{g} = \sum_{i=1}^K \alpha_i \cdot \bm{f}_i. 
\end{equation}
This can be regarded as a global pooling operation with adaptive spatial weight.
Figure~\ref{att} shows the details of the attention module. 

\subsection{Network architectures}\label{netarch}
The classification network (CNet) consists of a previously introduced feature extractor and an attention module, as well as a fully connected layer. Given an input image, the attention module generates its attended global feature and a coarse saliency map from the feature maps provided by the feature extractor. Then the fully connected layer transforms the global attended feature into a $C$-dimensional vector encoding the probability of each category, in which $C$ is the number of categories.

The architecture of the caption generation network (PNet) is similar to CNet. The main difference between them is that an LSTM layer replaces the fully connected layer of CNet. The LSTM layer takes the global attended feature as input and produces a sequence of $M$-dimensional vector, in which $M$ is the number of all candidate words. 
The saliency prediction network (SNet) is composed of a feature extractor; four dilated convolution layers with dilation rates 6, 12, 18, 24 respectively, and a deconvolution layer. The four dilated convolution layers take the feature map as input and predict four saliency maps. Then the four saliency maps are added together and upsampled to the input image size by the deconvolution layer. 

\subsection{Training with multiple supervision cues}\label{train}
Our training set $\mathcal{D}$ consists of three subsets: the classification dataset, the caption dataset and the unlabelled dataset. The classification dataset is denoted as $\mathcal{D}_c=\{(X^i, \bm{y}^i\}_{i=1}^{N_c}$, in which $y^i_j \in\{0,1\}, j=1, ..., C$ is the one-hot encoding of the categories appearing in image $X^i$. $N_c$ is the number of samples in $\mathcal{D}_c$. The caption dataset is denoted as $\mathcal{D}_p=\{(X^i, y^i_{1:T^i}\}_{i=1}^{N_p}$, in which $y^i_{1:T^i}$ is a sequence of $T^i$ words $(y^i_1, ..., y^i_{T^i})$. $N_p$ is the number of samples in $\mathcal{D}_p$. The unlabelled dataset is denoted as $\mathcal{D}_u = \{ X^i\}_{i=1}^{N_u}$, in which $N_u$ is the number of samples.

Given the input image $X$, CNet predicts the probability of the one-hot label of each category, denoted as $p(y_j|X), j=1, ..., C, y_j\in\{0,1\}$, and a saliency map $S_c$. Each element of $S_c$, denoted as $sc_i$, is the saliency score of the $i$-th region given by Equation~\ref{saliency_of_each_region}. PNet outputs the conditional distribution over candidate words at step $t$ of the sequence given the previous words $y_{1:t-1}$, denoted as $p(y_t|y_{1:t-1}, X), y_t=1, ..., M$. It also generates a saliency map $S_p$, of which each element is denoted as $sp_i$.

We define four loss functions to train the networks: a category localization loss $L_c$, a caption localization loss $L_p$, an attention transfer loss $L_{at}$ and an attention coherence loss $L_{ac}$. $L_c$ makes CNet find the most important regions for classification. $L_p$ makes PNet find the most important regions for generating caption. $L_{at}$ transfers supervision signal from the attention map of another network to the current network. $L_{ac}$ encourages two networks supervised by different sources to find the common salient regions. $L_c$ is defined as follow,
\begin{equation}\label{cls_loss}
L_c = -\frac{1}{N_c}\sum_{(X, \bm{y})\in \mathcal{D}_c}\left[ \sum_{j=1}^C \log p(y_j|X)
+ \beta \sum_{s\in S_c}\log (1-s) \right],
\end{equation}
where the first term is the log-likelihood, and the second term is the regularization that measures the cross-entropy between the saliency map $S_c$ and an all-zero map to prevent the trivial saliency map of having high responses at all locations. $\beta$ is a hyperparameter set to 0.005. Note that the saliency map $S_c$ and $S_p$ are generated for each input image and thus depend on the input image $X$.
Here and in the following equations, we omit this dependency of symbols for simplification. By minimize Equation~\ref{cls_loss}, CNet learns to predict the categories of the objects present in the input image. Meanwhile, the regularization term limits the amount of information flowing from the image regions to the classifier; therefore the network has to attend on the most important region, \ie, generate a reasonable saliency map, to predict the categories. 

The caption localization loss $L_p$ is defined as follow,
\begin{equation}\label{cap_loss}
\begin{split}
L_p &= -\frac{1}{N_p}\sum_{(X, y_{1:T})\in \mathcal{D}_p}\left[ \sum_{t=1}^T \log p(y_t|y_{1:t-1}, X)\right.\\
&+ \left. \beta \sum_{s\in S_p}\log (1-s) \right], 
\end{split}
\end{equation}
where the first term is the log likelihood and the second term is the regularization term as mentioned above. $\beta$ is set to 0.005. By minimizing Equation~\ref{cap_loss}, PNet learns to generate captions for the input image and find the salient regions corresponding to the caption. 

Constrained by is structures, CNet is unable to make direct use of the caption annotations, and PNet is unable to learn from the category annotations directly. To make full use of annotated data, we propose the attention transfer loss to let a network learn from the attention map of another network when its corresponding annotation is not available. 
Specifically, for an image annotated with category labels, we use the saliency map of CNet to select positive and negative samples (\ie salient regions and background regions) to supervise the saliency map of PNet. For an image annotated with captions, negative and positive samples are selected according to the saliency map of PNet to supervise the saliency map of CNet. Formally, the attention transfer loss is defined as follow,
\begin{equation}
\begin{split}
L_{at}&= -\frac{1}{N_c}\sum_{(X, \bm{y})\in \mathcal{D}_c}\left[\sum_{i\in I_c^+} \log sp_i + \sum_{i\in I_c^-} \log(1-sp_i)\right]\\
&-\frac{1}{N_p}\sum_{(X, y_{1:T})\in \mathcal{D}_p}\left[ \sum_{i\in I_p^+} \log sc_i + \sum_{i\in I_p^-} \log(1-sc_i) \right],
\end{split}
\end{equation}
where $I_c^+=\{i|sc_i\ge0.5 \}$ and $I_c^-=\{ i|sc_i<0.5\}$ are the indices of the salient and background regions selected according to the saliency map $S_c$. $I_p^+=\{i|sp_i\ge0.5 \}$ and $I_p^-=\{ i|sp_i<0.5\}$ are the indices of the salient and background regions selected according to $S_p$. 

For an input image, CNet and PNet respectively attend to the regions that are most important for predicting categories and generating captions. 
To make the networks find the generally salient regions, we incorporate low-level color similarity to refine the saliency maps of CNet and PNet and define an attention coherence loss on unlabelled data to match the saliency maps of CNet and PNet to the refined saliency map. 
Specifically, we segment each unlabelled image into superpixels using SLIC~\cite{achanta2010slic} and label the superpixels of which the saliency values are larger than the mean value in both $S_c$ and $S_p$ as salient seed, where the saliency value of a superpixel is defined as the average over its pixels. Then an affinity graph is constructed in which superpixels are nodes. Each superpixel is connected to its two-ring neighbors, and all superpixels on the image boundary are connected. 
The weight of the edge between the $m$-th and $n$-th nodes is defined by the Gaussian of the distance of the Lab color between the corresponding superpixels, \ie $w_{mn} = \exp(-|| c_m - c_n||/\sigma^2)$, in which $c_m, c_n$ denote the Lab color of the superpixel $m, n$, and $\sigma$ is set to 0.1. Inspired by~\cite{yang2013saliency}, we rank the color similarity of each superpixel with the salient seed by solving the following problem of ranking on data manifold~\cite{zhou2004ranking}:
\begin{equation}
\min_{\bm{h}} \frac{1}{2}(\sum_{m,n}w_{mn} ||\frac{h_m}{\sqrt{d_{mm}}}-\frac{h_n}{\sqrt{d_{nn}}}||^2+\mu\sum_{m}||h_m-z_m||^2),
\end{equation}
where $d_{mm}=\sum_n w_{mn}$. $\mu$ is set to 0.01. $z_m=1$ indicates the $m$-th superpixel is salient seed and $z_m=0$ otherwise. Let $D=\mbox{diag}\{d_{mm}\}$, the optimized $\bm{h}^*=(I-\gamma L)^{-1}\bm{z}$ is the ranking score of all superpixels, in which $L=D^{-1/2}WD^{-1/2}$ is the normalized Laplacian matrix and $\gamma=1/(1+\mu)$. We select the pixels of the superpixels whose ranking score is larger than the mean value of $\bm{h}^*$ as positive samples, denoted as $I_u^+$, and use other pixels as negative samples, denoted as $I_u^-$, to supervise the saliency maps of the two networks. The attention coherence loss is defined as follow,
\begin{equation}\label{L_ac}
\begin{split}
L_{ac}&= -\frac{1}{N_u}\sum_{X\in \mathcal{D}_u}\left[\sum_{i\in I_u^+} \log sc_i + \log sp_i \right.\\
&+\left. \sum_{i\in I_u^-} \log (1-sc_i) + \log (1-sp_i) \right].
\end{split}
\end{equation}

The loss function for training the whole system is given by the combination of the above four loss functions:
\begin{equation}
L = L_c + L_p + \lambda L_{at} + \lambda L_{ac},
\end{equation}
where $\lambda$ controls the weight of each term. We use a same weight $\lambda=0.01$ for $L_{at}$ and $L_{ac}$.

\subsection{Training the saliency prediction network}
Having trained CNet and PNet, we use their generated coarse saliency maps to train SNet. The two coarse saliency maps are averaged and resized to the original image size by bilinear interpolation. The averaged map is processed with CRF~\cite{krahenbuhl2011efficient} and then binarized into the pseudo labels. Let $Y$ be the pseudo labels, $S$ the output of SNet. We use the bootstrapping loss~\cite{reed2014training} to train SNet:
\begin{equation}
\begin{split}
L_b(S, Y) &= - \sum_i \left[ \delta y_i + (1-\delta)a_i \right] \log s_i \\
&+ \left[ \delta (1-y_i) + (1-\delta)(1-a_i)\right] \log (1-s_i), 
\end{split}
\end{equation}
where $y_i, s_i$ are the elements of $Y, S$ respectively, and $a_i = 1$ if $s_i\ge0.5$ else $a_i=0$. $\delta$ is set to 0.05. Note that we use CRF only when generating pseudo labels to train SNet. When testing, the saliency maps is predicted in an end-to-end manner without any post-processing.

\section{Experiments}
\subsection{Datasets and evaluation metrics}
We evaluate our method on five benchmark datasets: ECSSD~\cite{yan2013hierarchical}, PASCAL-S~\cite{li2014secrets}, SOD~\cite{martin2001database}, MSRA5K~\cite{liu2011learning} and DUT-OMRON~\cite{yang2013saliency}. 
\textbf{ECSSD} contains 1000 natural images with multiple objects of different sizes collected from the Internet. \textbf{PASCAL-S} is from the validation set of PASCAL VOC2010~\cite{everingham2010pascal} segmentation challenge and contains 850 natural images. \textbf{SOD} has 300 images and was designed originally for image segmentation; Jiang~\etal~\cite{jiang2013salient} generated the pixel-wise annotations of salient objects. \textbf{MSRA5K} has 5,000 images with a variety of image contents. \textbf{DUT-OMRON} contains 5,168 challenging images with one or more salient objects on complex backgrounds. 

We use Precision-Recall curve, mean absolute error (MAE) and maximum F-measure (max $F_\beta$ with $\beta^2$ set to 0.3 as suggested in~\cite{achanta2009frequency}) to quantitatively evaluate the performance of the proposed method and compare with other methods. 



\subsection{Implementation details}
We implement our method using Python with the PyTorch\footnote{https://github.com/pytorch} toolbox. Our code will be released for future comparisons\footnote{https://github.com/zengxianyu/mws}\footnote{http://ice.dlut.edu.cn/lu/}. In the first training stage, we train CNet and PNet on ImageNet detection dataset for multi-label classification and Microsoft COCO caption dataset as well as about 300,000 images from the ImageNet classification dataset as unlabelled data. In this training stage, we use the Adam optimizer~\cite{kingma2014adam} with batch size 36 and learning rate 0.0001. In the second training stage, we use the images of DUTS training set~\cite{wang2017learning} as unlabelled data and use the trained CNet and PNet to generate pseudo ground-truth to train SNet. In this training stage, we use the Adam optimizer with batch size 26 and learning rate 0.0001. All training images are resized to $256\times 256$. During training, we randomly crop and flip the images to avoid overfitting. When testing, the proposed method runs at about 103 fps with $256\times 256$ resolution on our computer with a 3.2GHz CPU, 32GB RAM and two GTX 1080Ti GPUs.

\subsection{Ablation studies}
\begin{figure}[t]
\begin{center}
 \includegraphics[width=\linewidth]{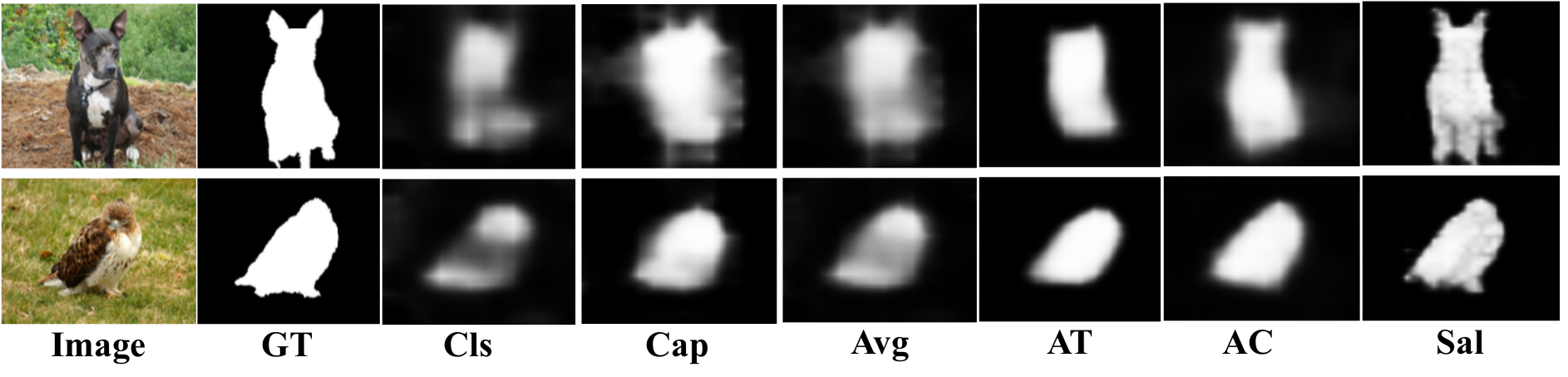}
\end{center}
  \caption{Visual effect of each component. \textbf{Image}: input image. \textbf{GT}: ground truth. \textbf{Cls}: the result of CNet trained with category localization loss $L_c$. \textbf{Cap}: the result of PNet trained with caption localization loss $L_p$. \textbf{Avg}: the average result of \textbf{Cls} and \textbf{Cap}. \textbf{AT}: Jointly training the two networks using the attention transfer loss $L_{at}$. \textbf{AC}: Jointly training the two networks and using unlabelled data for regularization. Loss on the unlabelled data is the attention coherence loss $L_{ac}$. \textbf{Sal}: Training SNet with the pseudo labels generated by CNet and PNet. }
\label{vis_steps}
\end{figure}

In this section, we analyze the contribution of each component including CNet, PNet, the attention transfer loss, the attention coherence loss (applied on unlabelled data), and SNet. The effect of each component in terms of maximum F-measure is shown in Table~\ref{tab_ablation}. The visual effect of each component is shown in Figure~\ref{vis_steps}. 

\noindent
\textbf{Learning from single supervision source}. We train CNet and PNet separately to explore the effect of each supervision source. Specifically, CNet is trained with image-level category labels using the category localization loss $L_c$ and PNet is trained with image captions using the caption localization loss $L_p$. Then we evaluate the performance of each network and the average results of the two networks. As shown in the 1-2 rows of Table~\ref{tab_ablation}, both CNet and PNet alone are not able to provide satisfactory results. The average result (the third row of Table~\ref{tab_ablation}) is better than each of the two, which demonstrates that the two supervision sources are complementary. 

\noindent
\textbf{Multi-source supervision with attention transfer loss}. \\Although averaging the results of CNet and PNet can improve the performance, the improvement is minimal. This is because information of training data is not used to the full by training the two networks separately and simply averaging the results. In contrast, by incorporating the attention transfer loss and jointly training the two networks, CNet benefits from the captions and PNet also benefits from the category labels. As a result, jointly training the two networks with attention transfer loss achieves a much better performance  (the fourth row of Table~\ref{tab_ablation}) than simply averaging the results (the third row of Table~\ref{tab_ablation}). 

\noindent
\textbf{Contribution of the unlabelled data}. To verify the contribution of unlabelled data, we train CNet and PNet jointly and using unlabelled data with the attention coherence loss. For labelled data, the loss is the sum of the category (or caption) localization loss and the attention transfer loss. For unlabelled data, we compute the attention coherence loss $L_{ac}$ as in Equation~\ref{L_ac}. The attention coherence loss encourages the networks to attend on more generally salient objects rather than the task-specific regions. As shown in the fifth row of Table~\ref{tab_ablation}, the performance is improved by incorporating unlabelled data and the attention coherence loss. 

\noindent
\textbf{Effect of the saliency prediction network}. After jointly training CNet and PNet with category labels, captions and unlabelled data, we use them to generate pseudo labels to train SNet. The performance of SNet is shown in the last row of Table~\ref{tab_ablation}. 
\begin{table}
\caption{Effect of each component in terms of maximum F-measure on ECSSD dataset. \textbf{Cls}: Training CNet using category localization loss $L_c$. \textbf{Cap}: Training PNet using caption localization loss $L_p$. \textbf{AT}: Jointly training the two networks with the attention transfer loss $L_{at}$. \textbf{AC}: Jointly training the two networks and using unlabelled data for regularization. Loss on the unlabelled data is the attention coherence loss $L_{ac}$. \textbf{Sal}: Training SNet with the pseudo labels generated by CNet and PNet. }
\label{tab_ablation}
\begin{center}
\begin{tabular}{|l|l|l|l|l|c|}
\hline
\textbf{Cls} & \textbf{Cap} & \textbf{AT} &\textbf{AC}& \textbf{Sal} & max $F_\beta$\\
\hline\hline
$\surd$ &         &         &         &         &0.720\\
        & $\surd$ &         &         &         &0.730\\
$\surd$ & $\surd$ &         &         &         &0.762\\
$\surd$ & $\surd$ & $\surd$ &         &         &0.786\\
$\surd$ & $\surd$ & $\surd$ & $\surd$ &         & 0.820\\
$\surd$ & $\surd$ & $\surd$ & $\surd$ &  $\surd$& 0.878\\
\hline
\end{tabular}
\end{center}
\vspace{-20pt}
\end{table}

\subsection{Performance comparison}
\begin{figure*}[t!]
\begin{center}
\includegraphics[width=\linewidth]{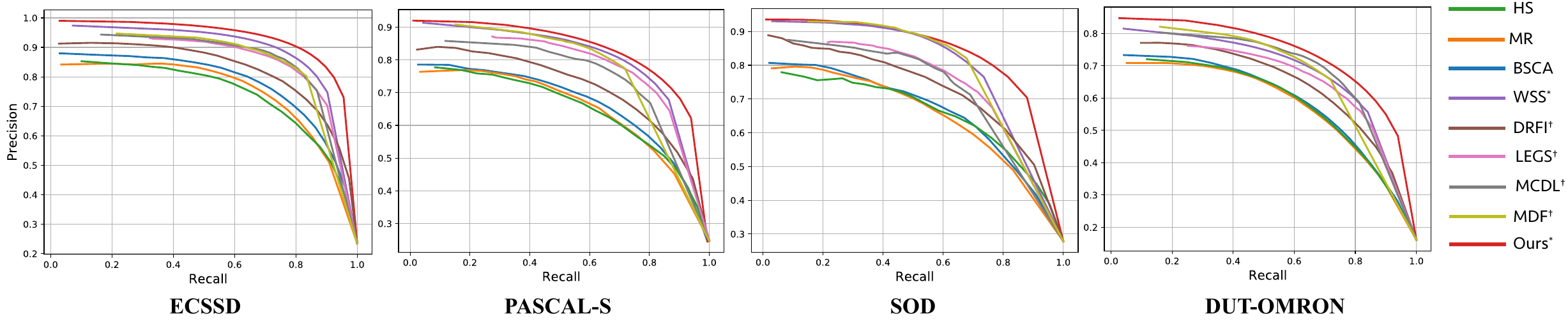}
\end{center}
  \caption{Precision-Recall curves. Our method outperforms unsupervised methods, weakly supervised method (marked with $^*$) and supervised methods (marked with $^\dag$). }
\label{pr_curve}
\end{figure*}

\begin{figure}[t!]
\begin{center}
\includegraphics[width=\linewidth]{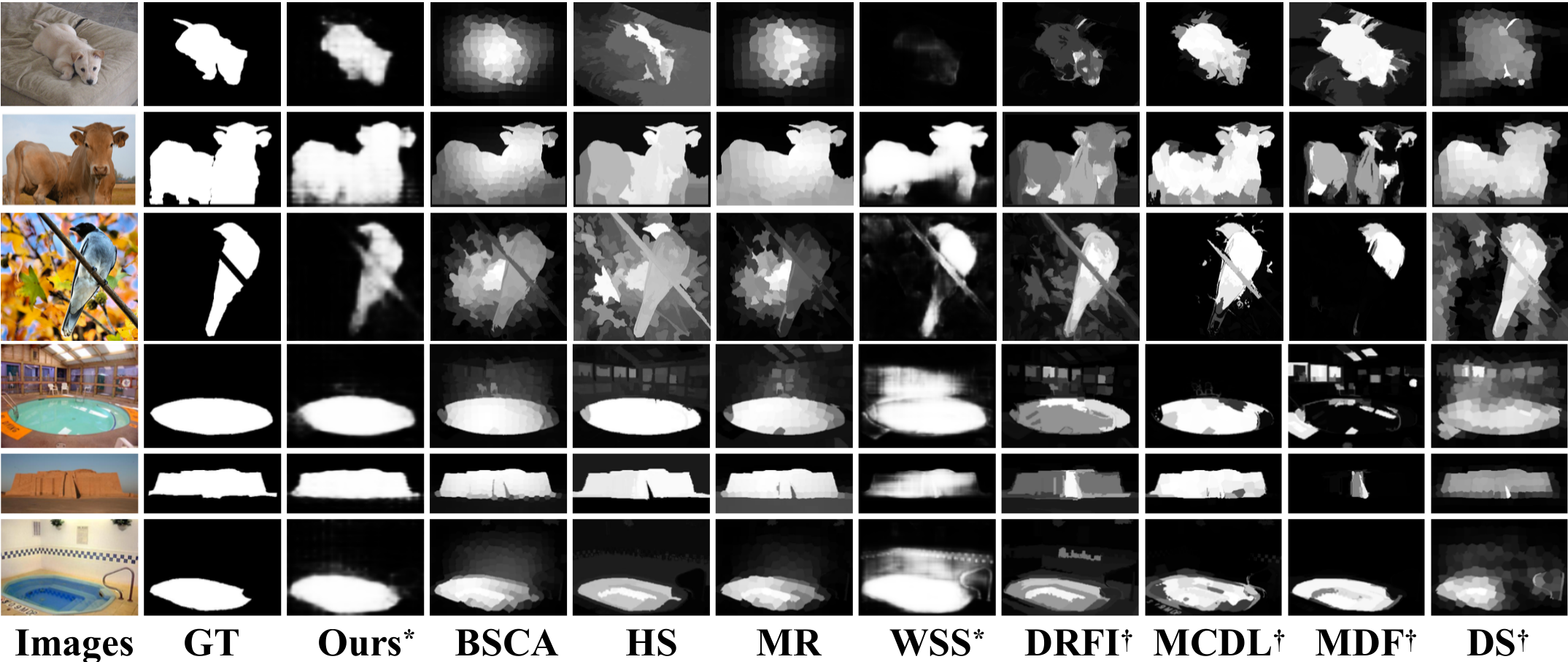}
\end{center}
  \caption{Visual comparison. Weakly and fully supervised methods are marked with $^*$ and $^\dag$ respectively.}
\label{visual}
\end{figure}

We compare the performance of our method and 11 stage-of-the-art methods, including five unsupervised methods BSCA~\cite{qin2015saliency}, MB+~\cite{zhang2015minimum}, MST~\cite{tu2016real}, MR~\cite{yang2013saliency}, HS~\cite{yan2013hierarchical}, one weakly supervised method WSS~\cite{wang2017learning}, and five fully supervised methods DRFI~\cite{jiang2013salient}, LEGS~\cite{wang2015deep}, MCDL~\cite{zhao2015saliency}, MDF~\cite{li2016visual}, DS~\cite{li2016deepsaliency}. The weakly supervised method WSS is trained with category labels of ImageNet detection dataset. The fully supervised methods DRFI, LEGS, MCDL, MDF and DS are trained with pixel-level saliency annotations. Except for DRFI, all the compared supervised methods are based on deep CNNs. We use the saliency maps provided by the authors or obtained by running the code provided by the authors for a fair comparison. The Precision-Recall curves (Figure~\ref{pr_curve}) and the score comparison (Table~\ref{score_weak}) show that our method outperforms all unsupervised methods with a large margin. As can be seen in Figure~\ref{pr_curve} and Table~\ref{score_weak}, the performance of our method is also better than another weakly supervised method WSS.
Figure~\ref{pr_curve} and Table~\ref{score_full} show that our method achieves comparable even better performance against fully supervised methods. As can be seen in Figure~\ref{pr_curve}, our method has a larger recall with the same precision. Table~\ref{score_full} shows  that our method outperforms fully supervised methods DRFI and LEGS. Our method also has better performance than MCDL, MDF and DS on most datasets. Visual comparison in Figure~\ref{visual} also demonstrates the superiority of our method. Compared with unsupervised methods, our methods can detect semantically salient objects that of low contrast to the background, \eg the dog in the first row, and the salient object in the cluttered background \eg the bird in the third row. Compared with another weakly supervised method WSS trained with only object categories, our method can better highlight the non-object salient regions such as water in the fourth and sixth row. 

\begin{table*}[t!]
\caption{\small Comparison with weakly supervised (marked with $^*$ and unsupervised methods in terms of maximum F-measure (the larger the better) and MAE (the smaller the better). The best scores are in bold. }
\label{score_weak}
\small
\footnotesize
  \renewcommand{\arraystretch}{1}
  \renewcommand{\tabcolsep}{3.5mm}
\centering
\begin{tabular}{|l|l|l|l|l|l|l|l|l|l|l|l|}
\hline
 &\multicolumn{2}{|c|}{{\bf ECSSD}} &\multicolumn{2}{|c|}{{\bf PASCAL-S}} &\multicolumn{2}{|c|}{{\bf SOD}} &\multicolumn{2}{|c|}{{\bf MSRA5K}} &\multicolumn{2}{|c|}{{\bf DUT-OMRON}}\\ 
\hline
\hline
Methods &max $F_\beta$&MAE &max $F_\beta$&MAE &max $F_\beta$&MAE &max $F_\beta$&MAE &max $F_\beta$&MAE \\
\hline
BSCA& 0.758&0.182& 0.663&0.223 &0.656&0.252 &0.829&0.132 &0.613&0.196 \\
\hline
MB+ & 0.736 &0.193 &0.673 &0.228 &0.658&0.255 &0.822&0.133 &0.621&0.193 \\
\hline
MST &0.724&0.155 &0.657&0.194 &0.647&0.223 &0.809&0.098 &0.588&0.161 \\
\hline
MR& 0.742&0.186& 0.650&0.232& 0.644&0.261& 0.821&0.128& 0.608&0.194\\
\hline
HS& 0.726&0.227& 0.644&0.264& 0.647&0.283& 0.815&0.162& 0.613&0.233\\
\hline
WSS$^*$& 0.856&0.104& 0.778&0.141& 0.780&0.170& 0.877&0.076& 0.687&0.118\\
\hline
Ours$^*$& \textbf{0.878}&\textbf{0.096}& \textbf{0.790}&\textbf{0.134}& \textbf{0.799}&\textbf{0.167}& \textbf{0.890}&\textbf{0.071}& \textbf{0.718}&\textbf{0.114}\\
\hline
\end{tabular}
\small
\end{table*}
\begin{table*}[t!]
\caption{\small Comparison with fully supervised methods in terms of maximum F-measure (the larger the better) and MAE (the smaller the better). Weakly supervised method is marked with $^*$ . MSRA5K dataset is absent as most supervised methods use it for training.}
\label{score_full}
\small
\footnotesize
  \renewcommand{\arraystretch}{1}
  \renewcommand{\tabcolsep}{3.5mm}
  
\centering
\begin{tabular}{|l|l|l|l|l|l|l|l|l|l|}
\hline
 &\multicolumn{2}{|c|}{{\bf ECSSD}} &\multicolumn{2}{|c|}{{\bf PASCAL-S}} &\multicolumn{2}{|c|}{{\bf SOD}} &\multicolumn{2}{|c|}{{\bf DUT-OMRON}}\\ 
\hline
\hline
Methods &max $F_\beta$&MAE &max $F_\beta$&MAE &max $F_\beta$&MAE &max $F_\beta$&MAE\\
\hline
DRFI&0.785&0.164           & 0.697&0.207          & 0.701&0.224         &0.651&0.145\\
\hline
LEGS& 0.827&0.118          & 0.761&0.155          & 0.733&0.196         & 0.671&0.140\\
\hline
MCDL& 0.837&0.101          & 0.743&0.145          & 0.730&0.181         & 0.703&\textbf{0.096}\\
\hline
MDF& 0.831&0.105           &0.768&0.146           &0.786&\textbf{0.159}          & 0.693&0.100\\
\hline
DS& \textbf{0.882}&0.122       & 0.763&0.176          & 0.784&0.190         &\textbf{0.739}&0.127\\
\hline
Ours$^*$& 0.878&\textbf{0.096}&\textbf{0.790}&0.134& \textbf{0.799}&0.167&0.718&0.114\\
\hline
\end{tabular}
\small
\end{table*}

\section{Conclusion and future work}
We propose a unified framework to train saliency detection models with diverse weak supervision sources. We use category labels, captions, and unlabelled data for training. We design A classification network (CNet) and a caption generation network (PNet), which learn from category labels and captions to generate saliency maps, respectively. An attention transfer loss is designed to transmit supervision signal between networks, such that the network purposed for one supervision source can benefit from another source. An attention coherence loss is defined on unlabelled data to encourage the networks to detect generally salient regions instead of task-specific regions. Final saliency predictions are made by a saliency prediction network (SNet) trained with pseudo labels generated by CNet and PNet. Experiments demonstrate the superiority of the proposed method, of which the performance compares favourably against unsupervised and weakly supervised methods, and is even better than some supervised methods.

The proposed framework is flexible and can be easily extended to integrate more supervision sources. Possible future directions include incorporating more supervision sources such as bounding box supervision, scribble supervision, and noisy saliency maps generated by unsupervised methods. It also can be extended to simultaneously exploit weak supervision sources, unlabelled data, and pixel-level annotations for semi-supervised learning.

\section*{Acknowledgements}
This work was supported by the National Natural Science Foundation of China (\#61725202, \#61829102 and \#61751212)



{\small
\bibliographystyle{ieee}
\bibliography{egbib}
}

\end{document}